\pdfoutput=1

\documentclass[11pt]{article}

\usepackage{EMNLP2023}

\usepackage{times}
\usepackage{latexsym}

\usepackage[T1]{fontenc}

\usepackage[utf8]{inputenc}

\usepackage{microtype}

\usepackage{inconsolata}

\usepackage{microtype}
\usepackage{graphicx}
\usepackage{amsmath}
\usepackage{caption}
\usepackage{subcaption}
\usepackage{algorithmic}
\usepackage{comment}
\usepackage{multirow}
\usepackage{url}

\title{BSpell: A CNN-Blended BERT Based Bangla Spell Checker}

\author{Chowdhury Rafeed Rahman \\
  National University of Singapore \\
  \texttt{e0823054@u.nus.edu}\And
  MD. Hasibur Rahman \\
  United International University \\
  \AND
  Samiha Zakir \and Mohammed Rafsan \\
  University of Texas Rio Grande Valley \\\And
   Mohammed Eunus Ali \\
  Bangladesh University of Engineering \\ and Technology
  }

\begin{document}
\maketitle
\begin{abstract}
Bangla typing is mostly performed using English keyboard and can be highly erroneous due to the presence of compound and similarly pronounced letters. Spelling correction of a misspelled word requires understanding of word typing pattern as well as the context of the word usage. A specialized BERT model named \textit{BSpell} has been proposed in this paper targeted towards word for word correction in sentence level. \textit{BSpell} contains an end-to-end trainable CNN sub-model named \textit{SemanticNet} along with specialized auxiliary loss. This allows \textit{BSpell} to specialize in highly inflected Bangla vocabulary in the presence of spelling errors. Furthermore, a hybrid pretraining scheme has been proposed for \textit{BSpell} that combines word level and character level masking. Comparison on two Bangla and one Hindi spelling correction dataset shows the superiority of our proposed approach. \textit{BSpell} is available as a Bangla spell checking tool via GitHub: \textit{https://github.com/Hasiburshanto/Bangla-Spell-Checker}.
\end{abstract}

\section{Introduction}
Bangla is the native language of 228 million people which makes it the sixth most spoken language in the world \footnote{https://www.babbel.com/en/magazine/the-10-most-spoken-languages-in-the-world}. This Sanskrit originated language has 11 vowels, 39 consonants, 11 modified vowels and 170 compound characters \cite{error1}. There is vast difference between Bangla grapheme representation and phonetic utterance for many commonly used words. As a result, fast typing of Bangla yields frequent spelling mistakes. Almost all Bangla native speakers type using English QWERTY layout keyboard \citep{qwerty} which makes it difficult to type Bangla compound characters, phonetically similar single characters and similar pronounced modified vowels correctly. Thus Bangla typing speed, if error-free typing is desired, is slow. An accurate spell checker (SC) can be a solution to this problem. 

Existing Bangla SCs include phonetic rule \citep{rel7,rel8} and clustering based methods \citep{rel9}. These methods do not take misspelled word context into consideration. Another N-gram based Bangla SC \citep{rel10} takes only short range previous context into consideration. Recent state-of-the-art (SOTA) spell checkers have been developed for Chinese language, where a character level confusion set (similar characters) guided sequence to sequence (seq2seq) model has been proposed by \citet{rel1}. Another research used similarity mapping graph convolutional network in order to guide BERT based character by character parallel correction \citep{rel4}. Both these methods require external knowledge and assumption about confusing character pairs existing in the language. The most recent Chinese SC offers an assumption free BERT architecture where error detection network based soft-masking is included \citep{rel3}. This model takes all $N$ characters of a sentence as input and produces the correct version of these $N$ characters as output in a parallel manner.   

\begin{figure}[!htb]
\centering
  \includegraphics[width=0.50\textwidth]{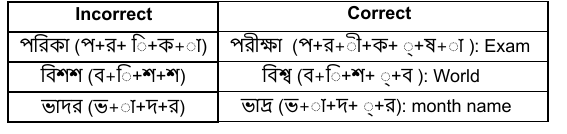}
  \caption{Heterogeneous character number between error word and corresponding correctly spelled word}
  \label{fig:hetero}
\end{figure}

One of the limitations in developing Bangla SC using SOTA BERT based implementation \citep{rel3} is that number of input and output characters in BERT has to be exactly the same. Such scheme is only capable of correcting substitution type errors. As compound characters are common in Bangla words, an error made due to the substitution of such characters also changes word length (see the table in Figure \ref{fig:hetero}). So, we introduce word level prediction in our proposed BERT based model.

\begin{figure}[!htb]
\centering
  \includegraphics[width=0.5\textwidth]{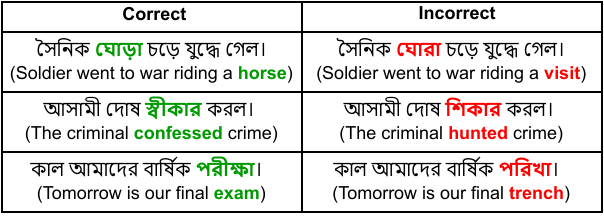}
  \caption{ample words that are correctly spelled accidentally, but are context-wise incorrect.}
  \label{fig:context}
\end{figure}

The table shown in Figure \ref{fig:context} illustrates the importance of context in Bangla SC. Although the red marked words of this figure are the misspelled versions of the corresponding green marked correct words, these red words are valid Bangla words. But if we check these red words based on sentence semantic context, we can realize that these words have been produced accidentally because of spelling error. An effective SC has to consider word pattern, its prior context and its post context.

\begin{figure}[!htb]
\centering
  \includegraphics[width=0.4\textwidth]{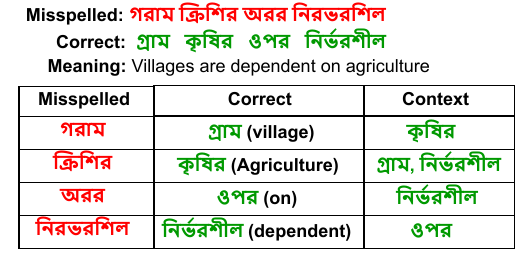}
  \caption{Necessity of understanding existing erroneous words for spelling correction of misspelled words}
  \label{fig:manyError}
\end{figure}

Spelling errors often span up to multiple words in a sentence. Figure \ref{fig:manyError} provides an example where all four words have been misspelled. The correction of each word has context dependency on a few other words of the same sentence. The problem is that these words that form the correction context are also misspelled. The table in the figure shows the words to look at in order to correct each misspelled word. In the original sentence (colored in red), all these words that need to be looked at for context are misspelled. If a SC cannot understand the approximate underlying meaning of these misspelled words, then we lose all context for correcting each misspelled word which is undesirable.

We propose a word level BERT \citep{BERT1} based model \textit{BSpell}. This model is capable of learning prior and post context dependency through the use of multi-head attention mechanism of stacked Transformer encoders \citep{BERT4}. The model uses CNN based learnable \textit{SemanticNet} sub-model to capture semantic meaning of both correct and misspelled words. \textit{BSpell} also uses specialized auxiliary loss to facilitate word level pattern learning and vanishing gradient problem removal. We introduce \textit{hybrid pretraining}for \textit{BSpell} to capture both context and word error pattern. We perform detailed evaluation on three error datasets that include a real life Bangla error dataset. Our evaluation includes detailed analysis on possible LSTM based SCs, SC variants of BERT and existing classic Bangla SCs.

\section{Related Works}
Several studies on Bangla SC development have been conducted in spite of Bangla being a low resource language. A phonetic encoding oriented Bangla word level SC based on Soundex algorithm was proposed by \citet{rel7}. This encoding scheme was later modified to develop a Double Metaphone encoding based Bangla SC \citep{rel8}. They took into account major context-sensitive rules and consonant clusters while performing their encoding scheme. Another word level Bangla SC able to handle both typographical and phonetic errors was proposed by \citet{rel9}. An N gram model was proposed by \citet{rel10} for checking sentence level Bangla word correctness. An encoder-decoder based seq2seq model was proposed by \citet{rel11} for Bangla sentence correction task which involved bad arrangement of words and missing words, though this work did not include incorrect spelling. A recent study has included Hindi and Telugu SC development, where mistakes are assumed to be made at character level \citep{rel2}. They have used attention based encoder-decoder modeling as their approach.

SOTA research in this domain involves Chinese SCs as it is an error prone language due to its confusing word segmentation, phonetically and visually similar but semantically different characters. A seq2seq model assisted by a pointer network was employed for character level spell checking where the network is guided by externally generated character confusion set \citep{rel1}. Another research
incorporated phonological and visual similarity knowledge of Chinese characters
into BERT based SC model by utilizing graph convolutional network \citep{rel4}. A recent BERT based SC has taken advantage of GRU (Gated Recurrent Unit) based soft masking mechanism and has achieved SOTA performance in Chinese character level SC in spite of not providing any external knowledge to the network \citep{rel3}. Another external knowledge free approach namely FASPell used BERT based seq2seq model \citep{relo3}. HanSpeller++ is notable among initially implemented Chinese SCs \citep{relo4}. It was an unified framework utilizing a hidden Markov model.

\section{Our Approach}
 \subsection{Problem Statement}
Suppose, an input sentence consists of $n$ words – $Word_1$, $Word_2$, …, $Word_n$. For each $Word_i$, we have to predict the right spelling, if $Word_i$ exists in the top-word list of our corpus. If $Word_i$ is a rare word (Proper Noun in most cases), we predict $UNK$ token denoting that we do not make any correction to such words. For correcting a particular $Word_i$ in a paragraph, we only consider other words of the same sentence for context information.

\subsection{BSpell Architecture} \label{model}

\begin{figure*}[!htb]
\centering
  \includegraphics[width=1.0\textwidth]{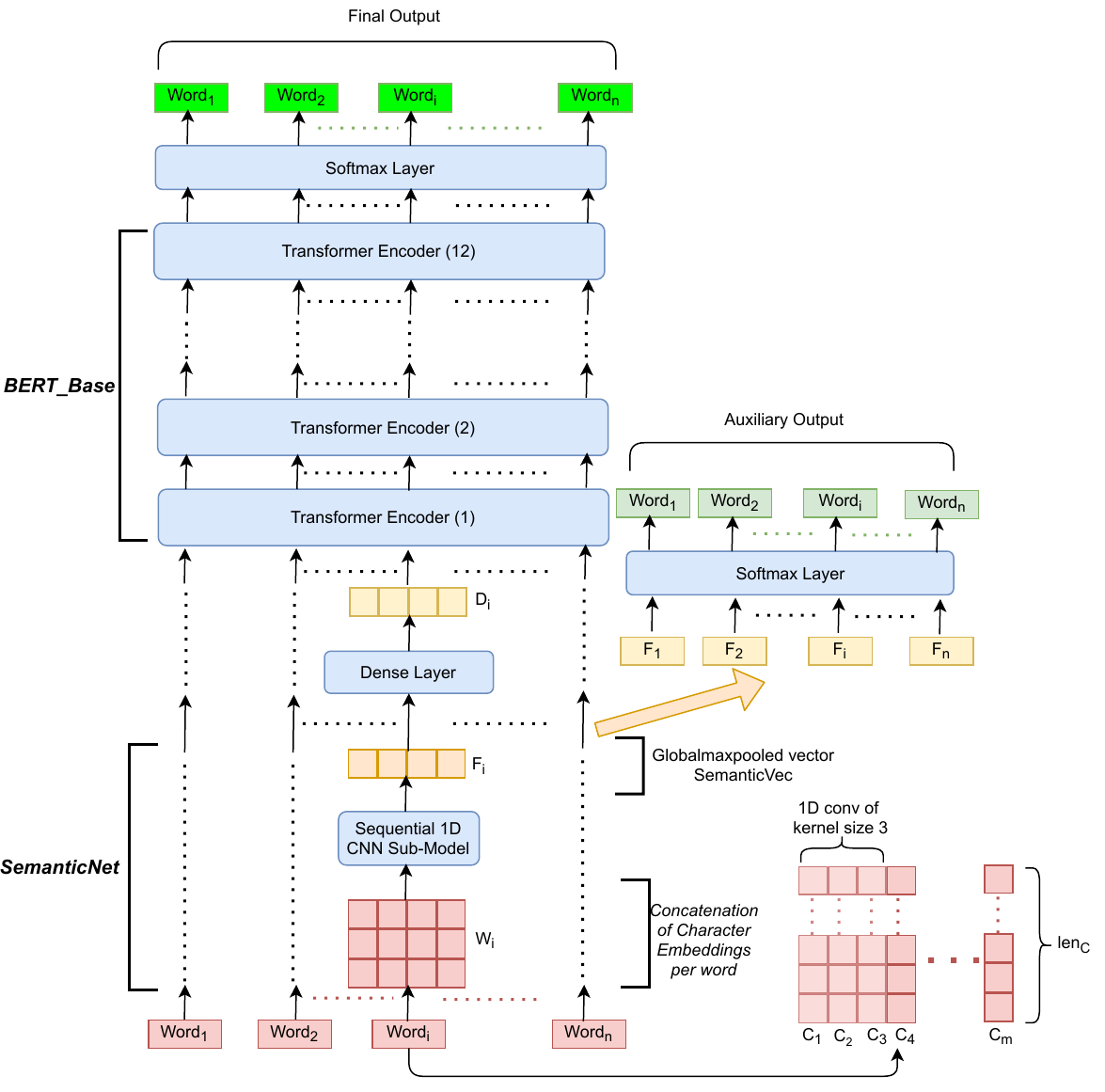}
  \caption{BSpell architecture details}
  \label{fig:BERT}
\end{figure*}

Figure \ref{fig:BERT} shows the details of \textit{BSpell} architecture. Each input word of the sentence is passed through the \textbf{\textit{SemanticNet} sub-model}. This sub-model returns us with a \textit{SemanticVec} vector representation for each input word. These vectors are then passed onto two separate branches (\textbf{main branch} and \textbf{secondary branch}) simultaneously. The main branch is similar to BERT\_Base architecture \citep{BERT5}. This branch provides us with the $n$ correct words corresponding to the $n$ input sentence words at its output side. The secondary branch consists of an output dense layer. This branch is used for the sole purpose of imposing \textbf{auxiliary loss} to facilitate \textit{SemanticNet} sub-model learning of misspelled word patterns.  

\subsubsection{SemanticNet Sub-Model} \label{semanticNet}
 Correcting a particular word requires the understanding of other relevant words in the same sentence. Unfortunately, those relevant words may also be misspelled. As humans, we can understand the meaning of a word even if it is misspelled because of our deep understanding at word syllable level and our knowledge of usual spelling error pattern. We want our model to have similar semantic level understanding of the words. We propose \textit{SemanticNet}, a sequential 1D CNN sub-model that is employed at each individual word level with a view to learning intra word syllable pattern. Details of individual word representation has been shown in the bottom right corner of Figure \ref{fig:BERT}. We represent each input word by a matrix (each character represented as a one hot vector).
We apply global max pooling on the final convolution layer output feature matrix of \textit{SemanticNet} which gives us the \textit{SemanticVec} vector representation of the input word. We get a similar \textit{SemanticVec} representation from each of our input words by independently applying the same \textit{SemanticNet} sub-model on each of their matrix representations. 

\subsubsection{BERT\_Base as Main Branch}
Each of the \textit{SemanticVec} vector representations obtained from the input words are passed parallelly on to our first Transformer encoder. 12 such Transformer encoders are stacked on top of each other. Each Transformer employs multi head attention mechanism, layer normalization and dense layer specific modification on each input vector. The attention mechanism applied on the word feature vectors in each transformer layer helps the words of the input sentence interact with one another extracting sentence context. We pass the final Transformer layer output vectors to a dense layer with Softmax activation function applied on each vector in an independent manner. So, now we have $n$ probability vectors from $n$ words of the input sentence. Each probability vector contains $len_P$ values, where $len_P$ is one more than the total number of top words considered (the additional word represents rare words). The top word corresponding to the index of the maximum probability value of $i^{th}$ probability vector represents the correct word for $Word_i$ of the input sentence.
 
 \subsubsection{Auxiliary Loss in Secondary Branch}
 Gradient vanishing problem is a common phenomena in deep neural networks, where weights of the shallow layers are not updated sufficiently during backpropagation. With the presence of 12 Transformer encoders on top of the \textit{SemanticNet} sub-model, the layers of this sub-model certainly lie in a shallow position. Although \textit{SemanticNet} constitutes a small initial portion of \textit{BSpell}, this portion is responsible for word pattern learning, an important task of SC. In order to eliminate gradient vanishing problem of \textit{SemanticNet} and to turn it into an effective pattern based word level spell checker, we introduce an auxiliary loss based secondary branch in \textit{BSpell}.
 Each of the $n$ \textit{SemanticVecs} obtained from the $n$ input words are passed parallelly on to a Softmax layer without any further modification. The outputs obtained from this branch are probability vectors similar to the main branch output. The total loss of \textit{BSpell} can be expressed as: $L_{Total} = L_{Final} + \lambda \times L_{Auxiliary}$. We want our final loss to have greater impact on model weight update as it is associated with the final prediction made by \textit{BSpell}. Hence, we impose the constraint $0 < \lambda < 1$. This secondary branch of \textit{BSpell} does not have any Transformer encoders through which the input words can interact to produce context information. The prediction made from this branch is dependent solely on misspelled word pattern extracted by \textit{SemanticNet}. This enables \textit{SemanticNet} to learn more meaningful word representation.

\subsection{BERT Hybrid Pretraining}

\begin{figure}[!htb]
    \begin{center}
    \includegraphics[width=1.0\columnwidth]{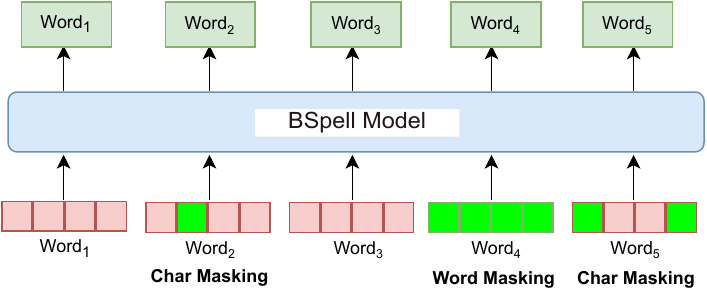}
    \end{center}
\caption{BERT hybrid pretraining}
\label{fig:hybrid}
\end{figure}

In contemporary BERT pretraining methods, each input word $Word_i$ maybe kept intact or maybe replaced by a default mask word in a probabilistic manner \citep{BERT1,BERT2}. BERT has to predict the masked words. Mistakes from the BERT side will contribute to loss value accelerating backpropagation based weight update. In this process, BERT learns to fill in the gaps, which in turn teaches the model language context. \citet{mask_pretrain} proposed incremental ways of pretraining the model for new NLP tasks. We take a more task specific approach for masking. In SC, recognizing noisy word pattern is important. But there is no provision for that in contemporary pretraining schemes and so, we propose hybrid masking (see Figure \ref{fig:hybrid}). Among $n$ input words in a sentence, we randomly replace $n_W$ words with a mask word $Mask_W$. Among the remaining $n - n_W$ words, we choose $n_C$ words for character masking. We choose $m_C$ characters at random from a word having $m$ characters to be replaced by a mask character $Mask_C$ during character masking. Such masked characters introduce noise in words and helps BERT to understand the probable semantic meaning of noisy/ misspelled words.

\section{Experimental Setup}

\subsection{Implemented Pretraining Schemes} \label{pretrain_detail}
We have experimented with three types of masking based pretraining schemes. During \textbf{word masking} we randomly select 15\% words of a sentence and replace those with a fixed mask word. During
\textbf{character masking}, we randomly select 50\% words of a sentence. For each selected word, we randomly mask 30\% of its characters by replacing each of them with a special mask character. Finally, during
\textbf{hybrid masking}, we randomly select 15\% words of a sentence and replace them with a fixed mask word. We randomly select 40\% words from the remaining words. For these selected words, we randomly mask 25\% of their characters.

\subsection{Dataset Specification}

\begin{table*}[!htb]
\centering
\begin{tabular}{|c|c|c|c|c|c|c|c|}
\hline
\textbf{Datasets}                                                                      & \textbf{\begin{tabular}[c]{@{}c@{}}Unique\\ Word\end{tabular}} & \textbf{\begin{tabular}[c]{@{}c@{}}Unique\\ Char\end{tabular}} & \textbf{\begin{tabular}[c]{@{}c@{}}Top\\ Word\end{tabular}} & \textbf{\begin{tabular}[c]{@{}c@{}}Train\\ Sample\end{tabular}} & \textbf{\begin{tabular}[c]{@{}c@{}}Validation\\ Sample\end{tabular}} & \textbf{\begin{tabular}[c]{@{}c@{}}Unique \\ Error Word\end{tabular}} & \textbf{\begin{tabular}[c]{@{}c@{}}Error \\ Word\\ Percentage\end{tabular}} \\ \hline
\begin{tabular}[c]{@{}c@{}}Prothom-Alo \\ Bangla\\ Synthetic Error\end{tabular}        & 262 K                                                          & 73                                                             & 35 K                                                        & 1 M                                                             & 200 K                                                                & 450 K                                                                 & 52\%                                                                        \\ \hline
\begin{tabular}[c]{@{}c@{}}Bangla Real \\ Error\end{tabular}                           & 14.5 K                                                         & 73                                                             & \_                                                          & 4.3 K                                                           & 2 K                                                                  & 10 K                                                                  & 36\%                                                                        \\ \hline
\begin{tabular}[c]{@{}c@{}}Bangla Pretrain \\ Corpus\end{tabular}                      & 513 K                                                          & 73                                                             & 40 K                                                        & 5.5 M                                                           & \_                                                                   & \_                                                                    & \_                                                                          \\ \hline
\begin{tabular}[c]{@{}c@{}}Hindi Synthetic \\ Error Corpus\\ (ToolsForIL)\end{tabular} & 20.5 K                                                         & 77                                                             & 15 K                                                        & 75 K                                                            & 16 K                                                                 & 5 K                                                                   & 10\%                                                                        \\ \hline
\begin{tabular}[c]{@{}c@{}}Hindi Pretrain \\ Corpus\end{tabular}                       & 370 K                                                          & 77                                                             & 40 K                                                        & 5.5 M                                                           & \_                                                                   & \_                                                                    & \_                                                                          \\ \hline
\end{tabular}
\caption{Dataset specification details}
  \label{tab:dataset}
\end{table*}

We have used one Bangla and one Hindi corpus with over 5 million (5 M) sentences for BERT pretraining (see Table \ref{tab:dataset}). Bangla pretraining corpus consists of Prothom Alo \footnote{https://www.prothomalo.com/} articles dated from 2014-2017 and BDnews24 \footnote{https://bangla.bdnews24.com/} articles dated from 2015-2017. The Hindi pretraining corpus consists of Hindi Oscar Corpus \footnote{https://www.kaggle.com/abhishek/hindi-oscar-corpus}, preprocessed Wikipedia articles \footnote{https://www.kaggle.com/disisbig/hindi-wikipedia-articles-172k}, HindiEnCorp05 dataset \footnote{http://hdl.handle.net/11858/00-097C-0000-0023-625F-0} and WMT Hindi News Crawl data \footnote{https://www.aclweb.org/anthology/W19-5301} (all of these are publicly available corpus).
We have used Prothom-Alo 2017 online newspaper dataset for Bangla SC training and validation purpose. Our errors in this corpus have been produced synthetically using the probabilistic algorithm described by \citet{error1}. We further validate our baselines and proposed methods on Hindi open source SC dataset, namely ToolsForIL \citep{rel2}. For real error dataset, we have collected a total of 6300 sentences from Nayadiganta \footnote{https://www.dailynayadiganta.com/} online newspaper. Then we have distributed the dataset among ten participants. They have typed (in regular speed) each correct sentence using English QWERTY keyboard producing natural spelling errors. It has taken 40 days to finish the labeling. Top words have been taken such that they cover at least 95\% of the corresponding corpus.

\subsection{\textit{BSpell} Architecture Hyperparameters}

\textit{SemanticNet} sub-model of \textit{BSpell} consists of a character level embedding layer producing a 40 size vector from each character, then 5 consecutive layers each consisting of 1D convolution (batch normalization and Relu activation in between each pair of convolution layers) and finally, a 1D global max pooling in order to obtain \textit{SemanticVec} representation from each input word. The five 1D convolution layers consist of $(64, 2), (64, 3), (128, 3), (128, 3), (256, 4)$ convolution, respectively. The first and second element of each tuple denote number of convolution filters and kernel size, respectively. We provide a weight of 0.3 ($\lambda$ value of loss function) to the auxiliary loss. The main branch of \textit{BSpell} is similar to BERT\_Base \citep{BERT5} in terms of stacking 12 Transformer encoders. Attention outputs from each Transformer is passed through a dropout layer \citep{drop} with a dropout rate of 0.3 and then layer normalized \citep{layer}.
We use \textit{Stochastic Gradient Descent (SGD)} Optimizer with a learning rate of 0.001 for our model weight update. We clip our gradient value and keep it below 5.0 to avoid gradient exploding problem.

\section{Results and Discussion}

\subsection{Training and Validation Details}
In case of Bangla SC, we randomly initialize the weights of model $M$. We use our large Bangla pretrain corpus for hybrid pretraining and get pretrained model $M_{pre}$. Next we split our benchmark synthetic spelling error dataset (Prothom-Alo) into 80\%-20\% training-validation set. We fine tune $M_{pre}$ using the 80\% training portion (obtaining fine tuned model $M_{fine}$) and report performance on the remaining 20\% validation portion. We use the Bangla real spelling error dataset in two ways - (1) We do not fine tune $M_{fine}$ on any of part of this data and use the entire dataset as an independent test set (result reported with the title \textit{real error (no fine tune)}) (2) We split this real error dataset into 80\%-20\% training-validation and fine tune $M_{fine}$ further using the 80\% portion, then validate on the remaining 20\% (result reported with the title \textit{real error (fine tuned)}).
In case of Hindi, the first two steps (pretraining and fine tuning) are the same. We have not constructed any real life spelling error dataset for Hindi. So, results are reported on the 20\% held out portion of the benchmark dataset.

\subsection{{BSpell} vs Contemporary BERT Variants} \label{BERT_comp}

\begin{table*}[!htb]
\centering
\begin{tabular}{|c|c|c|c|c|c|c|c|c|}
\hline

 & \multicolumn{2}{c|}{\textbf{\begin{tabular}[c]{@{}c@{}}Synthetic Error\\ (Prothom-Alo)\end{tabular}}} & \multicolumn{2}{c|}{\textbf{\begin{tabular}[c]{@{}c@{}}Real-Error\\ (No Fine Tune)\end{tabular}}} & \multicolumn{2}{c|}{\textbf{\begin{tabular}[c]{@{}c@{}}Real-Error\\ (Fine Tuned)\end{tabular}}} & \multicolumn{2}{c|}{\textbf{\begin{tabular}[c]{@{}c@{}}Synthetic Error\\ (Hindi)\end{tabular}}} \\ \cline{2-9} 
   
\multirow{-2}{*}{\textbf{\begin{tabular}[c]{@{}c@{}}Spell Checker\\ Architecture\end{tabular}}} 
            
             & ACC                                                & F1                                               & ACC                                              & F1                                             & ACC                                             & F1                                            & ACC                                             & F1                                            \\ \hline
BERT Seq2seq                                                                                    & 31.6\%                                             & 0.305                                            & 24.5\%                                           & 0.224                                          & 29.3\%                                          & 0.278                                         & 22.8\%                                          & 0.209                                         \\ \hline
BERT Base                                                                                       & 91.1\%                                             & 0.902                                            & 83\%                                             & 0.823                                          & 87.6\%                                          & 0.855                                         & 93.8\%                                          & 0.923                                         \\ \hline
Soft Masked BERT                                                                                & 92\%                                               & 0.919                                            & 84.2\%                                           & 0.832                                          & 88.1\%                                          & 0.862                                         & 94\%                                            & 0.933                                         \\ \hline
BSpell                                                                                          & \textbf{94.7\%}                                    & \textbf{0.934}                                   & \textbf{86.1\%}                                  & \textbf{0.859}                                 & \textbf{90.1\%}                                 & \textbf{0.898}                                & \textbf{96.2\%}                                 & \textbf{0.96}                                 \\ \hline
\end{tabular}
\caption{Comparing BERT based variants. Typical word masking based pretraining has been used on
all these variants. Real-Error (Fine Tuned) denotes fine tuning of the Bangla syn-
thetic error dataset trained model on real error dataset, while Real-Error (No Fine Tune) means directly
validating synthetic error dataset trained model on real error dataset without any further fine tuning.}
\label{tab:BERT}
\end{table*}

We start with \textbf{BERT Seq2seq} where the encoder and decoder portion consist of 12 stacked Transformers \citep{BERT1}. Predictions are made at character level. Similar architecture has been used in \textit{FASpell} \citep{relo3} for Chinese SC. A word is considered wrong if even one of its characters is predicted incorrectly. Hence character level seq2seq modeling achieves poor result (see Table \ref{tab:BERT}). Moreover, in most cases during sentence level spell checking, the correct spelling of the $i^{th}$ word of input sentence has to be the $i^{th}$ word in the output sentence as well. Such constraint is difficult to follow through such architecture design. \textbf{BERT Base} consisting of stacked Transformer encoders has two differences from the design proposed by \citet{rel4} - (i) We make predictions at word level instead of character level (ii) We do not incorporate any external knowledge about Bangla SC since such knowledge is not well established in the field. This approach achieves good performance in all four cases. \textbf{Soft Masked BERT} learns to apply specialized synthetic masking on error prone words in order to push the error correction performance of \textit{BERT Base} further. The error prone words are detected using a GRU sub-model and the whole architecture is trained end to end. Although \citet{rel3} implemented this architecture to make corrections at character level, our implementation does everything in word level. We have used popular FastText \citep{rep1} word representation for both \textit{BERT Base} and \textit{Soft Masked BERT}. \textbf{BSpell} shows decent performance improvement in all cases.  


\subsection{Comparing \textit{BSpell} Pretraining Schemes}

\begin{table*}[!htb]
\centering
\begin{tabular}{|c|c|c|c|c|c|c|c|c|}
\hline
 & \multicolumn{2}{c|}{\textbf{\begin{tabular}[c]{@{}c@{}}Synthetic Error\\ (Prothom-Alo)\end{tabular}}} & \multicolumn{2}{c|}{\textbf{\begin{tabular}[c]{@{}c@{}}Real-Error\\ (No Fine Tune)\end{tabular}}} & \multicolumn{2}{c|}{\textbf{\begin{tabular}[c]{@{}c@{}}Real-Error\\ (Fine Tuned)\end{tabular}}} & \multicolumn{2}{c|}{\textbf{\begin{tabular}[c]{@{}c@{}}Synthetic Error\\ (Hindi)\end{tabular}}} \\ \cline{2-9}

 \multirow{-2}{*}{\textbf{\begin{tabular}[c]{@{}c@{}}Pretraining\\ Scheme\end{tabular}}}
                                                                                      & ACC                                                & F1                                               & ACC                                              & F1                                             & ACC                                             & F1                                            & ACC                                             & F1                                            \\ \hline
Word Masking                                                                           & 94.7\%                                             & 0.934                                            & 86.1\%                                           & 0.859                                          & 90.1\%                                          & 0.898                                         & 96.2\%                                          & 0.96                                          \\ \hline
Character Masking                                                                      & 95.6\%                                             & 0.952                                            & 85.3\%                                           & 0.851                                          & 89.2\%                                          & 0.889                                         & 96.4\%                                          & 0.963                                         \\ \hline
Hybrid Masking                                                                         & \textbf{97.6\%}                                    & \textbf{0.971}                                   & \textbf{87.8\%}                                  & \textbf{0.873}                                 & \textbf{91.5\%}                                 & \textbf{0.911}                                & \textbf{97.2\%}                                 & \textbf{0.97}                                 \\ \hline
\end{tabular}
\caption{Comparing \textit{BSpell} exposed to various pretraining schemes}
 \label{tab:pretrain}
\end{table*}

We have implemented three different pretraining schemes (details provided in Subsection \ref{pretrain_detail}) on \textit{BSpell} before fine tuning on spell checker dataset.  
\textbf{Word masking} teaches \textit{BSpell} context of a language through a fill in the gaps sort of approach. SC is not all about filling in the gaps. It is also about what the writer wants to say, i.e. being able to predict a word even if some of its characters are blank (masked). \textbf{Character masking} takes a more drastic approach by completely eliminating the fill in the gap task. This approach masks a few of the characters residing in some of the input words of the sentence and asks \textit{BSpell} to predict these noisy words' original correct version. The lack of context in such pretraining scheme puts negative effect on performance over real error dataset experiments, where harsh errors exist and context is the only feasible way of correcting such errors (see Table \ref{tab:pretrain}). \textbf{Hybrid masking} focuses both on filling in word gaps and on filling in character gaps through prediction of correct word and helps \textit{BSpell} achieve SOTA performance.  

\subsection{\textit{BSpell} vs Possible LSTM Variants}

\begin{table*}[!htb]
\centering
\begin{tabular}{|c|cc|cc|cc|cc|}
\hline
\multirow{2}{*}{\textbf{\begin{tabular}[c]{@{}c@{}}Spell Checker\\ Architecture\end{tabular}}} & \multicolumn{2}{c|}{\textbf{\begin{tabular}[c]{@{}c@{}}Synthetic Error\\ (Prothom-Alo)\end{tabular}}} & \multicolumn{2}{c|}{\textbf{\begin{tabular}[c]{@{}c@{}}Real-Error\\ (No Fine Tune)\end{tabular}}} & \multicolumn{2}{c|}{\textbf{\begin{tabular}[c]{@{}c@{}}Real-Error\\ (Fine Tuned)\end{tabular}}} & \multicolumn{2}{c|}{\textbf{\begin{tabular}[c]{@{}c@{}}Synthetic Error\\ (Hindi)\end{tabular}}} \\ \cline{2-9} 
                                                                                               & \multicolumn{1}{c|}{ACC}                                     & F1                                     & \multicolumn{1}{c|}{ACC}                                   & F1                                   & \multicolumn{1}{c|}{ACC}                                  & F1                                  & \multicolumn{1}{c|}{ACC}                                   & F1                                 \\ \hline
BiLSTM                                                                                         & \multicolumn{1}{c|}{81.9\%}                                  & 0.818                                  & \multicolumn{1}{c|}{78.3\%}                                & 0.781                                & \multicolumn{1}{c|}{81.1\%}                               & 0.809                               & \multicolumn{1}{c|}{81.2\%}                                & 0.809                              \\ \hline
Stacked BiLSTM                                                                                 & \multicolumn{1}{c|}{83.5\%}                                  & 0.832                                  & \multicolumn{1}{c|}{80.1\%}                                & 0.80                                 & \multicolumn{1}{c|}{82.4\%}                               & 0.822                               & \multicolumn{1}{c|}{82.7\%}                                & 0.824                              \\ \hline
Attn\_Seq2seq (Char)                                                                           & \multicolumn{1}{c|}{20.5\%}                                  & 0.178                                  & \multicolumn{1}{c|}{15.4\%}                                & 0.129                                & \multicolumn{1}{c|}{17.3\%}                               & 0.152                               & \multicolumn{1}{c|}{22.7\%}                                & 0.216                              \\ \hline
BSpell                                                                                         & \multicolumn{1}{c|}{\textbf{97.6\%}}                         & \textbf{0.971}                         & \multicolumn{1}{c|}{\textbf{87.8\%}}                       & \textbf{0.873}                       & \multicolumn{1}{c|}{\textbf{91.5\%}}                      & \textbf{0.911}                      & \multicolumn{1}{c|}{\textbf{97.2\%}}                       & \textbf{0.97}                      \\ \hline
\end{tabular}
\caption{Comparing LSTM based variants with hybrid pretrained \textit{BSpell}. FastText word representation
has been used with LSTM portion of each architecture.}
  \label{tab:LSTM}
\end{table*}

\textbf{BiLSTM} is a many to many bidirectional LSTM (two layers) that takes in all $n$ words of a sentence at once and predicts their correct version as output \citep{LSTM1}. During SC, \textit{BiLSTM} takes in both previous and post context into consideration besides the writing pattern of each word and shows reasonable performance (see Table \ref{tab:LSTM}). In \textbf{Stacked BiLSTM}, we stack twelve many to many bidirectional LSTMs instead of just two. We see marginal improvement in SC performance in spite of such large increase in parameter number. \textbf{Attn\_Seq2seq} LSTM model utilizes attention mechanism at decoder side \citep{LSTM2}. This model takes in misspelled sentence characters as input and provides the correct sequence of characters as output \citep{rel2}. Due to word level spelling correction evaluation, this model faces the same problems as \textit{BERT Seq2seq} model discussed in Subsection \ref{BERT_comp}. Proposed \textbf{BSpell} outperforms these models by a large margin. 

\subsection{Ablation Study}

\begin{table*}[!htb]
\centering
\begin{tabular}{|c|cc|cc|cc|cc|}
\hline
\multirow{2}{*}{\textbf{\begin{tabular}[c]{@{}c@{}}BSpell\\ Variants\end{tabular}}} & \multicolumn{2}{c|}{\textbf{\begin{tabular}[c]{@{}c@{}}Synthetic Error\\ (Prothom-Alo)\end{tabular}}} & \multicolumn{2}{c|}{\textbf{\begin{tabular}[c]{@{}c@{}}Real-Error\\ (No Fine Tune)\end{tabular}}} & \multicolumn{2}{c|}{\textbf{\begin{tabular}[c]{@{}c@{}}Real-Error\\ (Fine Tuned)\end{tabular}}} & \multicolumn{2}{c|}{\textbf{\begin{tabular}[c]{@{}c@{}}Synthetic Error\\ (Hindi)\end{tabular}}} \\ \cline{2-9} 
                                                                                    & \multicolumn{1}{c|}{ACC}                                     & F1                                     & \multicolumn{1}{c|}{ACC}                                   & F1                                   & \multicolumn{1}{c|}{ACC}                                  & F1                                  & \multicolumn{1}{c|}{ACC}                                   & F1                                 \\ \hline
Original                                                                            & \multicolumn{1}{c|}{\textbf{97.6\%}}                         & \textbf{0.971}                         & \multicolumn{1}{c|}{\textbf{87.8\%}}                       & \textbf{0.873}                       & \multicolumn{1}{c|}{\textbf{91.5\%}}                      & \textbf{0.911}                      & \multicolumn{1}{c|}{\textbf{97.2\%}}                       & \textbf{0.97}                      \\ \hline
No Aux Loss                                                                         & \multicolumn{1}{c|}{96.3\%}                                  & 0.96                                   & \multicolumn{1}{c|}{86.9\%}                                & 0.865                                & \multicolumn{1}{c|}{90.5\%}                               & 0.90                                & \multicolumn{1}{c|}{95.4\%}                                & 0.949                              \\ \hline
No SemanticNet                                                                      & \multicolumn{1}{c|}{94.5\%}                                  & 0.94                                   & \multicolumn{1}{c|}{85.7\%}                                & 0.848                                & \multicolumn{1}{c|}{89.2\%}                               & 0.885                               & \multicolumn{1}{c|}{95.2\%}                                & 0.95                               \\ \hline
No Hybrid Pretrain                                                                  & \multicolumn{1}{c|}{94.7\%}                                  & 0.934                                  & \multicolumn{1}{c|}{86.1\%}                                & 0.859                                & \multicolumn{1}{c|}{90.1\%}                               & 0.898                               & \multicolumn{1}{c|}{96.2\%}                                & 0.96                               \\ \hline
\end{tabular}
\caption{Comparing \textit{BSpell} with its variants created by removing one of its novel features}
\label{tab:ablation}
\end{table*}

\textit{BSpell} has three unique features - (1) secondary branch with auxiliary loss (possible to remove this branch), (2) 1D CNN based SemanticNet sub-model (can be replaced by simple \textit{Byte Pair Encoding (BPE)} \citep{BERT4}) and (3) hybrid pretraining (can be replaced by word masking based pretraining). Table \ref{tab:ablation} demonstrates the results we obtain after removing any one of these features. In all cases, the results show a downward trend compared to the original architecture.        

\subsection{Existing Bangla Spell Checkers vs \textit{BSpell}}

\begin{table}[!htb]
\centering
\begin{tabular}{|c|cc|cc|}
\hline
\multirow{2}{*}{\textbf{\begin{tabular}[c]{@{}c@{}}Spell \\ Checker\end{tabular}}} & \multicolumn{2}{c|}{\textbf{\begin{tabular}[c]{@{}c@{}}Synthetic Error\\ (Prothom-Alo)\end{tabular}}} & \multicolumn{2}{c|}{\textbf{\begin{tabular}[c]{@{}c@{}}Real-Error\\ (No Fine Tune)\end{tabular}}} \\ \cline{2-5} 
                                                                                   & \multicolumn{1}{c|}{ACC}                                     & F1                                     & \multicolumn{1}{c|}{ACC}                                   & F1                                   \\ \hline
Phonetic                                                                           & \multicolumn{1}{c|}{61.2\%}                                  & 0.582                                  & \multicolumn{1}{c|}{43.5\%}                                & 0.401                                \\ \hline
Clustering                                                                         & \multicolumn{1}{c|}{52.3\%}                                  & 0.501                                  & \multicolumn{1}{c|}{44.2\%}                                & 0.412                                \\ \hline
BSpell                                                                             & \multicolumn{1}{c|}{\textbf{97.6\%}}                         & \textbf{0.971}                         & \multicolumn{1}{c|}{\textbf{87.8\%}}                       & \textbf{0.873}                       \\ \hline
\end{tabular}
\caption{Existing Bangla spell checkers vs \textit{BSpell}}
\label{tab:Existing}
\end{table}

\textit{Phonetic} rule based SC takes a Bangla phonetic rule based hard coded approach \citep{rel12}, where a hybrid of Soundex \citep{rel7} and Metaphone \citep{rel8} algorithm has been used. \textit{Clustering} based SC on the other hand follows some predefined rules on word cluster formation, distance measurement and correct word suggestion \citep{rel9}. Since these two SCs are not learning based, fine tuning is not applicable for them. They do not take misspelled word context into consideration while correcting that word. As a result, their performance is poor especially in Bangla real error dataset (see Table \ref{tab:Existing}). \textit{BSpell} outperforms these Bangla SCs by a wide margin.

\subsection{Is \textit{BSpell} Language Specific?}
\textit{BSpell} has originally been designed keeping the unique characteristics of Sanskrit originated languages such as Bangla and Hindi in mind. Here we see how this model performs on English which is very different from Bangla in terms of structure. We experiment on an English spelling error dataset published by \citet{English_data}. The training set consists of 1.6 million sentences. The authors created a confusion set consisting of 109K misspelled-correct word pairs for 17K popular English words. 20\% of the words of the training set have been converted to spelling error based on this confusion set. The authors created BEA-60K test set from BEA-2019 shared task consisting of natural English spelling errors. The best correction rate achieved by the authors was around 80\% using LSTM based ELMo model, whereas \textit{BSpell} has achieved a correction rate of 86.2\%. We have also experimented with \textit{BERT\_Base} model on this test set where we have used byte pair encoding as word representation. \textit{BERT\_Base} has achieved an error correction rate of 85.6\%. It is clear that \textit{BSpell} and \textit{BERT\_Base} do not have that much difference in performance when it comes to English compared to Bangla and Hindi. 

\subsection{Effectiveness of \textit{SemanticNet}}

\begin{figure}[!htb]
\centering
  \includegraphics[width=0.4\textwidth]{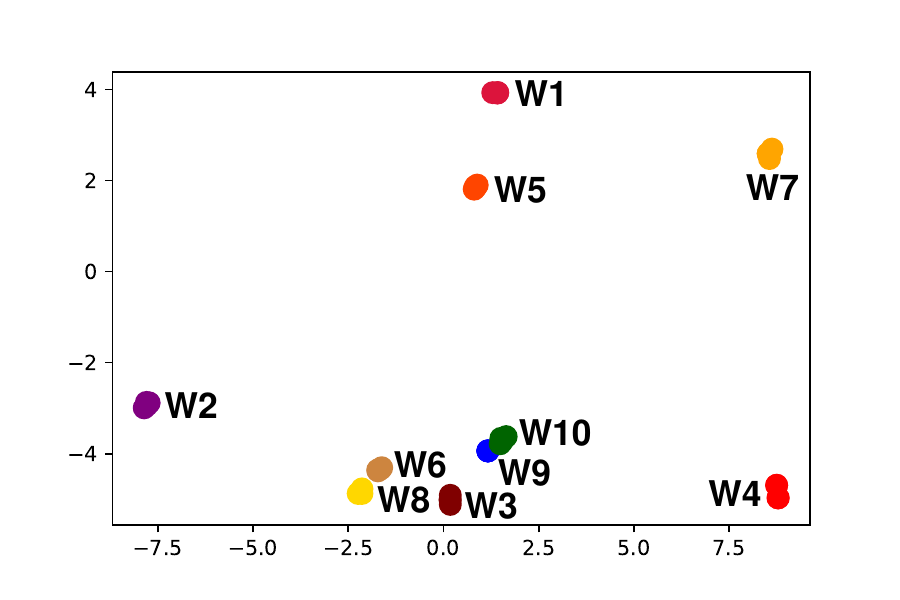}
  \caption{Visualizing \textit{SemanticVec} representation of 10 popular words with their error variants}
  \label{fig: err_vec}
\end{figure}

The main motivation behind the inclusion of \textit{SemanticNet} in \textit{BSpell} is to obtain vector representations of error words as close as possible to their corresponding correct words. We take 10 frequently occurring Bangla words and collect three real life error variations of each of these words. We produce \textit{SemanticVec} representation of all 40 of these words using \textit{SemanticNet}. We use principal component analysis (PCA) \citep{PCA} on each of these \textit{SemanticVecs} and plot them in two dimensions. Finally, we implement K-Means Clustering algorithm using careful initialization with $K=10$ \citep{Kmeans}. Figure \ref{fig: err_vec} shows the 10 clusters obtained from this algorithm. Each cluster consists of a popular word and its three error variations. In all cases, the correct word and its three error versions are so close in the graph plot that they almost form a single point.

\section{Conclusion}
In this paper, we have proposed a SC named \textit{BSpell} for Bangla and Hindi language. \textit{BSpell} uses \textit{SemanticVec} representation of input misspelled words and a specialized auxiliary loss for the enhancement of spelling correction performance. The model exploits the concept of hybrid masking based pretraining. We have also investigated into the limitations of existing Bangla SCs as well as other SOTA SCs proposed for high resource languages. \textit{BSpell} has two main limitations - (a) it cannot handle accidental merge or split of words and (b) it cannot correct misspelled rare words. A potential research direction can be to eradicate these limitations by designing models that can perform prediction at sub-word level which includes white space characters and punctuation marks.

\section{Limitations} \label{limit}
\textit{BSpell} model provides a word for word correction, i.e., number of input words and number of output words have to be exactly the same. Unfortunately, during accidental word merging or word splitting, number of input and output words differ and so in such cases \textit{BSpell} will fail in resolving such errors. This type of error is more common in Chinese language. The advantage for us is that this type of error is rare in Bangla and Hindi as the words of these languages are clearly spaced in sentences. So, people will rarely perform accidental merge or split of words. Another limitation is that \textit{BSpell} has been trained to correct only the top Bangla and Hindi words that cover 95\% of the entire corpus. As a result, this spell checker will face problems while correcting spelling errors in rare words. For such rare words, \textit{BSpell} simply provides \textit{UNK} as output which means that it is not sure what to do with these words. An advantage here is that most of these rare words are some form of proper nouns which should not be corrected and should ideally be left alone as they are. For example, someone may have an uncommon name. We do not want our model to correct that person's name to some commonly used name.\\
An immediate research direction is to overcome the limitations of the proposed method. A straightforward way of dealing with the word merge, word split and rare word correction problem is to model spelling errors at character level (sequence-to-sequence type approach). We have taken this trivial attempt and have failed miserably (see the performance reported in the first row of Table \ref{tab:BERT}). Solving these problems while maintaining the current spelling correction performance of \textit{BSpell} can be a challenge. Another interesting future direction is to investigate on personalized Bangla and Hindi spell checker which has the ability to take user personal preference and writing behaviour into account. The main challenge here is to effectively utilize user provided data that must be collected in an online setting. Recently, deep learning based automatic grammatical error correction has gained a lot of attention in English language \citep{grammar_correct1}, \citep{grammar_correct2}, \citep{grammar_correct3}. SOTA grammar correction models developed for English can be trained and tested on Bangla and Hindi spell checking tasks as part of future research effort. Such benchmarking studies can play a vital role in pushing the boundaries of low resource language correction automation.

\bibliography{main}

\begin{thebibliography}{30}
\expandafter\ifx\csname natexlab\endcsname\relax\def\natexlab#1{#1}\fi

\bibitem[{Athiwaratkun et~al.(2018)Athiwaratkun, Wilson, and Anandkumar}]{rep1}
Ben Athiwaratkun, Andrew~Gordon Wilson, and Anima Anandkumar. 2018.
\newblock Probabilistic fasttext for multi-sense word embeddings.
\newblock \emph{arXiv preprint arXiv:1806.02901}.

\bibitem[{Ba et~al.(2016)Ba, Kiros, and Hinton}]{layer}
Jimmy~Lei Ba, Jamie~Ryan Kiros, and Geoffrey~E Hinton. 2016.
\newblock Layer normalization.
\newblock \emph{arXiv preprint arXiv:1607.06450}.

\bibitem[{Bahdanau et~al.(2014)Bahdanau, Cho, and Bengio}]{LSTM2}
Dzmitry Bahdanau, Kyunghyun Cho, and Yoshua Bengio. 2014.
\newblock Neural machine translation by jointly learning to align and
  translate.
\newblock \emph{arXiv preprint arXiv:1409.0473}.

\bibitem[{Chen and Xia(2009)}]{Kmeans}
Zhang Chen and Shixiong Xia. 2009.
\newblock K-means clustering algorithm with improved initial center.
\newblock In \emph{2009 Second International Workshop on Knowledge Discovery
  and Data Mining}, pages 790--792. IEEE.

\bibitem[{Cheng et~al.(2020)Cheng, Xu, Chen, Jiang, Wang, Wang, Chu, and
  Qi}]{rel4}
Xingyi Cheng, Weidi Xu, Kunlong Chen, Shaohua Jiang, Feng Wang, Taifeng Wang,
  Wei Chu, and Yuan Qi. 2020.
\newblock Spellgcn: Incorporating phonological and visual similarities into
  language models for chinese spelling check.
\newblock \emph{arXiv preprint arXiv:2004.14166}.

\bibitem[{Chollampatt and Ng(2017)}]{grammar_correct2}
Shamil Chollampatt and Hwee~Tou Ng. 2017.
\newblock Connecting the dots: Towards human-level grammatical error
  correction.
\newblock In \emph{Proceedings of the 12th Workshop on Innovative Use of NLP
  for Building Educational Applications}, pages 327--333.

\bibitem[{Chollampatt and Ng(2018)}]{grammar_correct1}
Shamil Chollampatt and Hwee~Tou Ng. 2018.
\newblock A multilayer convolutional encoder-decoder neural network for
  grammatical error correction.
\newblock In \emph{Proceedings of the AAAI Conference on Artificial
  Intelligence}, volume~32.

\bibitem[{Devlin et~al.(2018)Devlin, Chang, Lee, and Toutanova}]{BERT1}
Jacob Devlin, Ming-Wei Chang, Kenton Lee, and Kristina Toutanova. 2018.
\newblock Bert: Pre-training of deep bidirectional transformers for language
  understanding.
\newblock \emph{arXiv preprint arXiv:1810.04805}.

\bibitem[{Etoori et~al.(2018)Etoori, Chinnakotla, and Mamidi}]{rel2}
Pravallika Etoori, Manoj Chinnakotla, and Radhika Mamidi. 2018.
\newblock \href {https://doi.org/10.18653/v1/P18-3021} {Automatic spelling
  correction for resource-scarce languages using deep learning}.
\newblock In \emph{Proceedings of {ACL} 2018, Student Research Workshop}, pages
  146--152, Melbourne, Australia. Association for Computational Linguistics.

\bibitem[{Gong et~al.(2019)Gong, He, Li, Qin, Wang, and Liu}]{BERT5}
Linyuan Gong, Di~He, Zhuohan Li, Tao Qin, Liwei Wang, and Tieyan Liu. 2019.
\newblock Efficient training of bert by progressively stacking.
\newblock In \emph{International Conference on Machine Learning}, pages
  2337--2346. PMLR.

\bibitem[{Hong et~al.(2019)Hong, Yu, He, Liu, and Liu}]{relo3}
Yuzhong Hong, Xianguo Yu, Neng He, Nan Liu, and Junhui Liu. 2019.
\newblock Faspell: A fast, adaptable, simple, powerful chinese spell checker
  based on dae-decoder paradigm.
\newblock In \emph{Proceedings of the 5th Workshop on Noisy User-generated Text
  (W-NUT 2019)}, pages 160--169.

\bibitem[{Islam et~al.(2018)Islam, Sarkar, Hussain, Hasan, Farid, and
  Shatabda}]{rel11}
Sadidul Islam, Mst~Farhana Sarkar, Towhid Hussain, Md~Mehedi Hasan, Dewan~Md
  Farid, and Swakkhar Shatabda. 2018.
\newblock Bangla sentence correction using deep neural network based sequence
  to sequence learning.
\newblock In \emph{2018 21st International Conference of Computer and
  Information Technology (ICCIT)}, pages 1--6. IEEE.

\bibitem[{Jayanthi et~al.(2020)Jayanthi, Pruthi, and Neubig}]{English_data}
Sai~Muralidhar Jayanthi, Danish Pruthi, and Graham Neubig. 2020.
\newblock Neuspell: A neural spelling correction toolkit.
\newblock \emph{arXiv preprint arXiv:2010.11085}.

\bibitem[{Khan et~al.(2014)Khan, Saha, Sarker, and Rahman}]{rel10}
Nur~Hossain Khan, Gonesh~Chandra Saha, Bappa Sarker, and Md~Habibur Rahman.
  2014.
\newblock Checking the correctness of bangla words using n-gram.
\newblock \emph{International Journal of Computer Application}, 89(11).

\bibitem[{Liu et~al.(2019)Liu, Ott, Goyal, Du, Joshi, Chen, Levy, Lewis,
  Zettlemoyer, and Stoyanov}]{BERT2}
Yinhan Liu, Myle Ott, Naman Goyal, Jingfei Du, Mandar Joshi, Danqi Chen, Omer
  Levy, Mike Lewis, Luke Zettlemoyer, and Veselin Stoyanov. 2019.
\newblock Roberta: A robustly optimized bert pretraining approach.
\newblock \emph{arXiv preprint arXiv:1907.11692}.

\bibitem[{Mandal and Hossain(2017)}]{rel9}
Prianka Mandal and BM~Mainul Hossain. 2017.
\newblock Clustering-based bangla spell checker.
\newblock In \emph{2017 IEEE International Conference on Imaging, Vision \&
  Pattern Recognition (icIVPR)}, pages 1--6. IEEE.

\bibitem[{Noyes(1983)}]{qwerty}
Jan Noyes. 1983.
\newblock The qwerty keyboard: A review.
\newblock \emph{International Journal of Man-Machine Studies}, 18(3):265--281.

\bibitem[{Saha et~al.(2019)Saha, Tabassum, Saha, and Akter}]{rel12}
Sourav Saha, Faria Tabassum, Kowshik Saha, and Marjana Akter. 2019.
\newblock \emph{BANGLA SPELL CHECKER AND SUGGESTION GENERATOR}.
\newblock Ph.D. thesis, United International University.

\bibitem[{Schuster and Paliwal(1997)}]{LSTM1}
Mike Schuster and Kuldip~K Paliwal. 1997.
\newblock Bidirectional recurrent neural networks.
\newblock \emph{IEEE transactions on Signal Processing}, 45(11):2673--2681.

\bibitem[{Shlens(2014)}]{PCA}
Jonathon Shlens. 2014.
\newblock A tutorial on principal component analysis.
\newblock \emph{arXiv preprint arXiv:1404.1100}.

\bibitem[{Sifat et~al.(2020)Sifat, Rahman, Rafsan, and Rahman}]{error1}
Md~Habibur~Rahman Sifat, Chowdhury~Rafeed Rahman, Mohammad Rafsan, and Hasibur
  Rahman. 2020.
\newblock Synthetic error dataset generation mimicking bengali writing pattern.
\newblock In \emph{2020 IEEE Region 10 Symposium (TENSYMP)}, pages 1363--1366.
  IEEE.

\bibitem[{Srivastava et~al.(2014)Srivastava, Hinton, Krizhevsky, Sutskever, and
  Salakhutdinov}]{drop}
Nitish Srivastava, Geoffrey Hinton, Alex Krizhevsky, Ilya Sutskever, and Ruslan
  Salakhutdinov. 2014.
\newblock Dropout: a simple way to prevent neural networks from overfitting.
\newblock \emph{The journal of machine learning research}, 15(1):1929--1958.

\bibitem[{Stahlberg and Kumar(2021)}]{grammar_correct3}
Felix Stahlberg and Shankar Kumar. 2021.
\newblock Synthetic data generation for grammatical error correction with
  tagged corruption models.
\newblock \emph{arXiv preprint arXiv:2105.13318}.

\bibitem[{Sun et~al.(2020)Sun, Wang, Li, Feng, Tian, Wu, and
  Wang}]{mask_pretrain}
Yu~Sun, Shuohuan Wang, Yukun Li, Shikun Feng, Hao Tian, Hua Wu, and Haifeng
  Wang. 2020.
\newblock Ernie 2.0: A continual pre-training framework for language
  understanding.
\newblock In \emph{Proceedings of the AAAI Conference on Artificial
  Intelligence}, volume~34, pages 8968--8975.

\bibitem[{UzZaman and Khan(2004)}]{rel7}
Naushad UzZaman and Mumit Khan. 2004.
\newblock A bangla phonetic encoding for better spelling suggesions.
\newblock Technical report, BRAC University.

\bibitem[{UzZaman and Khan(2005)}]{rel8}
Naushad UzZaman and Mumit Khan. 2005.
\newblock A double metaphone encoding for approximate name searching and
  matching in bangla.

\bibitem[{Vaswani et~al.(2017)Vaswani, Shazeer, Parmar, Uszkoreit, Jones,
  Gomez, Kaiser, and Polosukhin}]{BERT4}
Ashish Vaswani, Noam Shazeer, Niki Parmar, Jakob Uszkoreit, Llion Jones,
  Aidan~N Gomez, {\L}ukasz Kaiser, and Illia Polosukhin. 2017.
\newblock Attention is all you need.
\newblock In \emph{Advances in neural information processing systems}, pages
  5998--6008.

\bibitem[{Wang et~al.(2019)Wang, Tay, and Zhong}]{rel1}
Dingmin Wang, Yi~Tay, and Li~Zhong. 2019.
\newblock \href {https://doi.org/10.18653/v1/P19-1578} {Confusionset-guided
  pointer networks for {C}hinese spelling check}.
\newblock In \emph{Proceedings of the 57th Annual Meeting of the Association
  for Computational Linguistics}, pages 5780--5785, Florence, Italy.
  Association for Computational Linguistics.

\bibitem[{Xiong et~al.(2015)Xiong, Zhang, Zhang, Hou, and Cheng}]{relo4}
Jinhua Xiong, Qiao Zhang, Shuiyuan Zhang, Jianpeng Hou, and Xueqi Cheng. 2015.
\newblock Hanspeller: a unified framework for chinese spelling correction.
\newblock In \emph{International Journal of Computational Linguistics \&
  Chinese Language Processing, Volume 20, Number 1, June 2015-Special Issue on
  Chinese as a Foreign Language}.

\bibitem[{Zhang et~al.(2020)Zhang, Huang, Liu, and Li}]{rel3}
Shaohua Zhang, Haoran Huang, Jicong Liu, and Hang Li. 2020.
\newblock Spelling error correction with soft-masked bert.
\newblock \emph{arXiv preprint arXiv:2005.07421}.

\end{thebibliography}
\bibliographystyle{acl_natbib}




\end{document}